\documentclass[letterpaper, 10 pt, conference]{ieeeconf}  

\IEEEoverridecommandlockouts                              

\usepackage{graphics} 
\usepackage{epsfig} 
\usepackage{amsmath} 
\usepackage{amssymb}  
\usepackage{booktabs}
\usepackage{multirow}
\usepackage{makecell}
\usepackage{color}
\title{\LARGE \bf
Target-point Attention Transformer: A novel trajectory predict network for end-to-end autonomous driving
}

\author{Jingyu Du$^{1}$, Yang Zhao$^{*1}$ and Hong Cheng$^{1}$
\thanks{$^{1}$Jingyu Du, Yang Zhao, and Hong Cheng are with School of Automation Engineering, University of Electronic Science and Technology of China, 611731 Chengdu, China.}%
\thanks{$^{*}$ Corresponding Author: {\tt\small yzhao@uestc.edu.cn}}
}

\begin{document}

\maketitle
\thispagestyle{empty}
\pagestyle{empty}

\begin{abstract}

In the field of autonomous driving, there have been many excellent perception models for object detection, semantic segmentation, and other tasks, but how can we effectively use the perception models for vehicle planning? Traditional autonomous vehicle trajectory prediction methods not only need to obey traffic rules to avoid collisions, but also need to follow the prescribed route to reach the destination. In this paper, we propose a Transformer-based trajectory prediction network for end-to-end autonomous driving without rules called Target-point Attention Transformer network (TAT). We use the attention mechanism to realize the interaction between the predicted trajectory and the perception features as well as target-points. We demonstrate that our proposed method outperforms existing conditional imitation learning and GRU-based methods, significantly reducing the occurrence of accidents and improving route completion. We evaluate our approach in complex closed loop driving scenarios in cities using the CARLA simulator and achieve state-of-the-art performance.

\end{abstract}

\section{INTRODUCTION}

With the continuous development of deep learning, end-to-end autonomous driving is gradually becoming a hot topic\cite{Rhinehart2018DeepIM, Filos2020CanAV, Sadat2020PerceivePA, Cui2021LookOutDM, Xu2016EndtoEndLO}. End-to-end autonomous driving uses neural networks to directly map sensor inputs (cameras, LIDAR, IMU, etc.) to future trajectories\cite{Bojarski2020TheNP, Chen2019LearningBC, Prakash2021MultiModalFT, Chitta2021NEATNA, Jaeger2021, Chen2022LearningFA} or low-level control actions(e.g. throttle, brake and steering angle)\cite{Codevilla2017EndtoEndDV, Liang2018CIRLCI, Codevilla2019ExploringTL, OhnBar2020LearningSD, Chen2021LearningTD, Zhang2021EndtoEndUD, Chekroun2021GRIGR}, eliminating the need for complex rule base design. It is certainly exciting, as we know that rule bases are hardly designed to cover the full range of situations. Currently, the model for end-to-end autonomous driving can be thought of as an encoder-decoder structure\cite{Bansal2018ChauffeurNetLT, Chen2022LearningFA, Shao2022SafetyEnhancedAD}, where the encoder is responsible for encoding the information around the vehicle, and the decoder uses the encoded information for future trajectory or low-level control action prediction. For the encoders, there have been developments from the original ResNet networks\cite{Codevilla2017EndtoEndDV, Codevilla2019ExploringTL} to the multi-stage fusion of images and radar called Transfuser\cite{Prakash2021MultiModalFT}, and the interpretable network called Interfuser\cite{Shao2022SafetyEnhancedAD}. However, there have been few improvements for decoders, especially the method of future trajectory prediction.

   \begin{figure*}[thpb]
      \centering
      \includegraphics[scale=0.40]{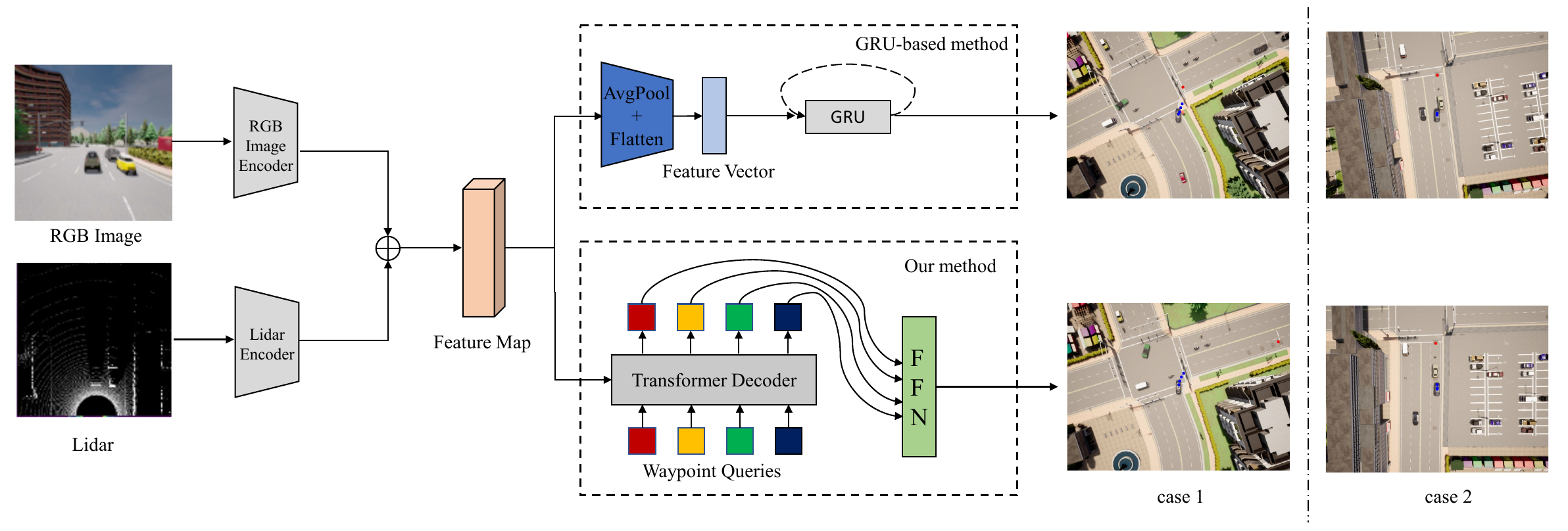}
      \caption{The GRU-based trajectory prediction network commonly adopt an AvgPool layer and a Flatten layer to convert 2D perception feature maps into 1D feature vectors. This dimensionality reduction can result in a loss of important perceptual features,, which may cause the autonomous vehicle to be unable to complete the route normally (${case\, 1}$ driving off the road, ${case\, 2}$ colliding with other vehicle, blue points are predicted waypoints, red points are target-points). Our method employs the Transformer's attention mechanism to directly interact with the 2D perceptual features and predict the future trajectory. Our method significantly reduces the occurrence of accidents and improves the completion of the route.}
      \label{figurelabel1}
   \end{figure*}
   
Unlike trajectory prediction tasks in other fields, for the autonomous driving task, the vehicle needs to make the correct decision at intersections, such as going straight, turning left, or turning right, based on advanced navigation commands(e.g. follow lane, turn right/left, change lane). Currently, there are two main types of decoder structures: conditional imitation learning network (CIL)\cite{Codevilla2017EndtoEndDV} and GRU-based\cite{Cho2014LearningPR} network. Codevilla ${et\,al.}$ proposed CIL structure, which trains multiple decoders to be responsible for the straight, left or right turn of the vehicle according to the advanced navigation commands. The disadvantages of this approach are obvious, firstly, training multiple networks for similar functions results in significant redundancy and waste of computational resources. In addition, this type of approach often performs poorly when passing curves due to dataset bias, making the completion of routes often low. Prakash ${et\,al.}$\cite{Prakash2021MultiModalFT} have proposed GRU-based methods that no longer use high-level navigation commands but instead use high-level target-points for vehicle navigation (these points are sparse and they can be hundreds of metres apart). However, passing the perception features from the encoder to the GRU cell usually requires a dimensionality reduction of the features, this dimensionality reduction can result in a loss of important perceptual features, which can lead to suboptimal predictions and potentially dangerous driving behavior, such as driving off the road or colliding with other vehicles, as shown in Figure \ref{figurelabel1}. The recent success of Transformer structures for pedestrian trajectory prediction\cite{Giuliari2020TransformerNF, Yu2020SpatioTemporalGT, Zhao2021SpatialChannelTN, Saleh2020PedestrianTP} demonstrates the potential of Transformer for trajectory prediction. Compared with RNN networks such as GRU, Transformer can input higher dimensional perception features and its attention mechanism is more conducive to smooth trajectory prediction. To this end, we propose the Target-point Attention Transformer network(TAT), witch uses the Transformer's attention mechanism to directly interact with the 2D perceptual features and predict the future trajectory. This approach preserves more of the relevant perceptual features and enables smoother, more accurate trajectory predictions. As a result, TAT significantly reduces the occurrence of accidents and improves the vehicle's ability to complete the route successfully. To the best of our knowledge, we are the first to propose using Transformer for end-to-end autonomous driving trajectory prediction.

The contributions can be summarized as follows:

(1) The proposed Target-point Attention Transformer model utilizes attention mechanism to predict the future trajectory of autonomous vehicles, which significantly reduces the occurrence of collisions and improves route completion. 

(2) Quantitative experiments are conducted on the CARLA, different trajectory prediction methods are compared and analyzed. Experiment results demonstrate the effectiveness of the proposed method.





The structure of this paper is as follows. In Section II, we present a comprehensive review of the related works concerning end-to-end driving models and trajectory forecasting using Transformer. In Section III, we describe our proposed method in detail. Section IV outlines the experimental setup, and Section V presents the results. Finally, we conclude this paper in Section VI. 

\section{Related Work}

\subsection{End-to-end Autonomous Driving}

With the continuous development of deep learning, learning-based end-to-end autonomous driving has become an active research topic. Currently, research in this field can be broadly categorized into two areas: reinforcement learning (RL) and imitation learning (IL). The RL methods in automatic driving is often combined with IL method. Liang ${et\,al.}$\cite{Liang2018CIRLCI} uses the supervision method to pretrain the model firstly, and then uses the Deep Deterministic Policy Gradient (DDPG) strategy for training. \cite{Toromanoff2019EndtoEndMR, Chekroun2021GRIGR, Zhao2022CADREAC} use the pretrained perception model to perform the perception task in RL. While RL has the potential to address the issue of dataset distribution shifts\cite{Chen2021LearningTD}, their training complexity often limits their practical applications.

Imitation learning realizes automatic driving by imitating the behavior of experts. The neural network is often trained under the supervision of experts' behaviors by collecting the data of cameras, laser radars and other sensors as input\cite{Bojarski2020TheNP, Chen2019LearningBC, Prakash2021MultiModalFT, Chitta2021NEATNA}. And the data of experts behaviors typically has two forms, actions and trajectories. Nvidia proposed PilotNet\cite{Bojarski2016EndTE}, which directly maps the image pixels of a single forward facing camera to the steering command via CNN. Codevilla ${et\,al.}$\cite{Codevilla2017EndtoEndDV} add a measurement encoder to fuse the vehicle status features, e.g. current speed and location. And it proposed conditional imitation learning with using multiple branches for different high level commands to drive to cross the intersection. Codevilla ${et\,al.}$\cite{Codevilla2019ExploringTL} proposed an improved method based on CIL, it use a speed prediction head to improve the inertia problem of end-to-end automatic driving\cite{Wen2020FightingCA}. The above methods directly predict actions. However, these methods usually can only predict the action of the current time step, which lead to the actions of vehicle more unstable and discontinuous. In addition, such methods are difficult for data augmentation. Wu ${et\,al.}$\cite{Wu2022TrajectoryguidedCP} presented a multi-task learning(MTL) method, it predicts the future trajectories first, and predict future actions through attention mechanism. It realized multi-step action prediction and achieved good results on the public CARLA Leaderboard\cite{CARLALearderboard}. 

For the trajectory methods, Chen ${et\,al.}$\cite{Chen2019LearningBC} introduced a knowledge distillation method. It utilizes privileged information (such as Bird's Eye View (BEV) map) to train a privileged agent, which generates a set of heatmaps. These heatmaps are passed through a soft-argmax layer (SA) to obtain waypoints for all commands. The sensorimotor agent is then trained using the privileged agent as a supervisor. Chitta ${et\,al.}$\cite{Chitta2021NEATNA} proposed a method that uses attention maps to compress high dimensional 2D image features into a more compact BEV representation for autonomous driving. It achieves this by using a series of intermediate attention maps that iteratively process the input image. NEAT also employs a waypoint offset prediction map to transform the discretely predicted waypoints into a dense prediction task. Transfuser\cite{Prakash2021MultiModalFT} use a multi-stage CNN-transformer to fuse the RGB image and LIDAR, then it uses a single GRU to auto-regress waypoints. Learning from All Vehicles (LAV)\cite{Chen2022LearningFA} also adopted a temporal GRU module to auto-regress waypoints. Note that such methods usually require a PID controller to convert waypoints into low-level control actions. And the trajectory-based methods have the advantage of providing smoother and more continuous driving behaviors than action-based methods. Additionally, trajectory-based methods can be more easily extended to incorporate additional constraints\cite{Shao2022SafetyEnhancedAD}, such as safety or efficiency requirements, by adjusting the trajectory planning algorithm.

\subsection{Transformer in trajectory forecasting}

Transformer first proposed in the field of Natural Language Processing\cite{Vaswani2017AttentionIA, Devlin2019BERTPO, Lan2019ALBERTAL, Young2017RecentTI, Yang2019XLNetGA}, and it quickly dominated this field with its unique attention mechanism. The ability of Transformer to parallelize computation and capture long-term dependencies has made it a popular choice for sequence modeling tasks. Recently the transformer architecture has also achieved success in the field of Computer Vision\cite{Dosovitskiy2020AnII, Carion2020EndtoEndOD, Gao2021FastCO, Liu2022PETRPE, Gabeur2020MultimodalTF} and trajectory prediction\cite{Giuliari2020TransformerNF, Yu2020SpatioTemporalGT}. Giuliari ${et\,al.}$\cite{Giuliari2020TransformerNF} use the basic transformer decoder architecture for pedestrian trajectory prediction, and achieve SOTA on multiple datasets. Compared to RNN, Transformer has the ability of parallel training, and can process higher dimensional features, these make the transformer performs better than RNN in the field of trajectory forecasting.

\section{Method}

In our work, we propose a novel waypoint prediction network for end-to-end autonomous driving. The following sections briefly introduce the Transfuser backbone and detail the design of our waypoint prediction network. 

\subsection{Problem Setting} 

The goal of whole network is to learn a policy $\pi$ so that the vehicle can reach the destination ${{u}_{G}}$ along the predefined route, $u_{1}^{G}\in ({{u}_{1}},...,{{u}_{g}},...,{{u}_{G}})$,  ${{u}_{g}}\in {{\mathbb{R}}^{2}}$ and can obey the traffic rules and avoid collisions with other traffic participants.

For the perception backbone, the goal is to encode the high dimensional observations of environment, $\mathsf{\mathcal{X}}$, to the lower perception features $\mathsf{\mathcal{F}}$. 
Then the waypoint prediction network takes perception features $\mathsf{\mathcal{F}}$ and navigation points $u$ in to generate the future waypoints $\mathcal{W}=\{{{w}_{i}}\}_{i=1}^{Z}$, $Z$ is the number of prediction waypoints.


The policy $\pi$ is trained in a supervised manner using the collected data $D$, with the loss function $\mathcal{L}$.	

\begin{equation}
\begin{aligned}
& \underset{\pi}{\mathop{argmin}}\,{{\mathbb{E}}_{(\mathcal{X},\mathcal{W},u)\sim \mathcal{D}}}[\mathcal{L}(\mathcal{W},\pi (\mathcal{X},u))]
\end{aligned}
\end{equation}

Assuming access to an inverse dynamics model implemented as a PID controller for low-level control, the system can generate the necessary steering, throttle, and brake commands given the predicted future waypoints $\mathcal{W}$. The actions can be determined as $a = \theta(\mathcal{W})$.

\textbf{Input Representation:} 
Following Tranfuser, our model takes an RGB front camera image and Lidar point cloud as input. For the image ${x}_{t}\in {{\mathbb{R}}^{H\times W\times C}}$ where $H=W=256$, $C=3$, we use the parameters on ImageNet for normalization. For the Lidar point cloud, it converted to a histogram pseudo-image ${v}_{t}\in {{\mathbb{R}}^{H\times W\times D}}$, where $D=2$. Since we consider the points within 32m in front of the ego-vehicle and 16m to each of the sides, each grid represents $0.125m\times 0.125m$ area. And we use the horizontal plane as the boundary to divide the points on the horizontal plane in the first dimension, and other divisions in the second dimension.

\textbf{Output Representation:} 
We predict the future waypoints $\mathcal{W}$ of the ego vehicle. $\mathcal{W}=\{{{w}_{i}}\}_{i=1}^{Z}$ is in the coordinate system with the ego vehicle as the origin, the front is the positive direction of the x-axis, and the left is the positive direction of the y-axis. Following \cite{Chitta2022TransFuserIW}, we take $Z=4$.

\subsection{Transfuser Backbone}

Transfuser use ResNet\cite{he2016deep} to process the image ${{x}_{t}}$ and the LIDAR BEV ${{v}_{t}}$. During processing, several Transformer modules are used to fuse the intermediate feature maps between both modalities. The Transformer module takes the intermediate features as token, the output is split into two parts and respectively sum with the existing feature maps. Note that the Transformer modules work at $8\times 8$ resolution, so there is a downsampling operation at the input and an upsampling operation at the output. For more details we refer to \cite{Prakash2021MultiModalFT, Chitta2022TransFuserIW}.

\begin{figure}[thpb]
      \centering
      \includegraphics[scale=0.22]{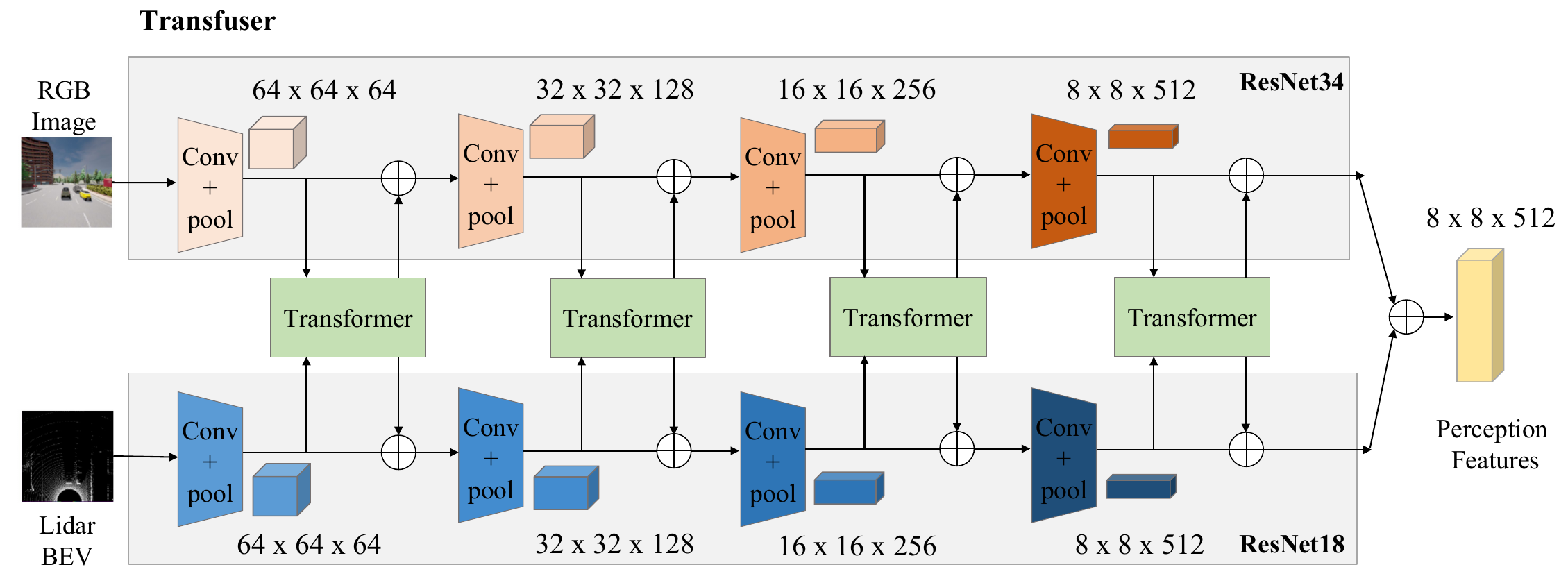}
      \caption{We discard the last avgpool layer and flatten layer of Transfuser to output full scale perception feature map. In addition, we cancel the speed input during fusion because it is one of the reason for the inertia problem\cite{Chitta2022TransFuserIW}}
      \label{figurelabel2}
   \end{figure}

The original Transfuser model uses the average pool layer and flatten layer to reduce the extracted 2D feature map to 1D feature vector so that the perception features can pass into GRU cell, this processing may lead to a large loss of information, which may cause the autonomous vehicle to be unable to complete the route normally. In the architecture illustrated in Figure \ref{figurelabel2}, we remove the final avgpool layer and flatten layer to produce a full-scale perception feature map. This map is then fed into our Transformer decoder-based waypoint prediction network, which is described in detail in the following section.

\subsection{Target-point Attention Waypoint Prediction Network}

\begin{figure}[thpb]
      \centering
      \includegraphics[scale=0.28]{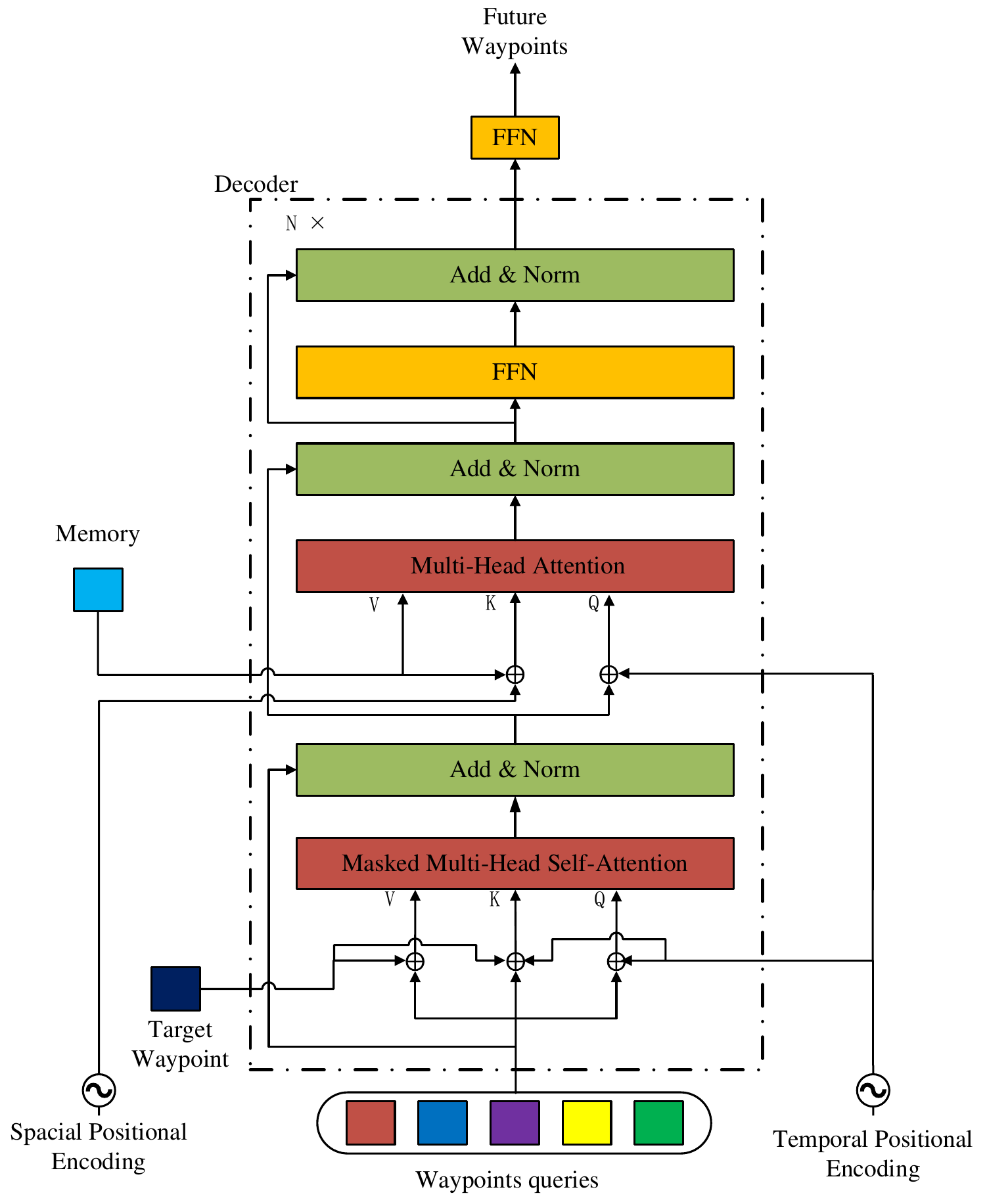}
      \caption{Architecture of TAT, We added the target-point vector during the calculation of self-attention, so that the self-driving vehicle can pass the intersection correctly. And the cross-attention mechanism of Transformer can complete the interaction between waypoints and perception features.}
      \label{figurelabel3}
   \end{figure}

Our model takes the full-scale perception features $\mathcal{F}\in {{\mathbb{R}}^{{{H}_{0}}\times {{W}_{0}}\times {{d}_{\operatorname{model}}}}}$ as the input to predict the future trajectory of the ego vehicle. Our key idea is to use Transformer's self-attention mechanism  to build dependency between waypoints and target-points, and use cross attention mechanism to build dependency between waypoints and perception features. The overall structure of the our network is shown in the Figure \ref{figurelabel3}.

\textbf{Self-Attention:}
As we need to predict a set $\mathcal{W}=\{{{w}_{i}}\in {{\mathbb{R}}^{2}}\}_{i=1}^{Z}$ waypoints, we use a set of waypoints queries $\mathcal{Q}_{l}=\{{{e}_{wi}}\in {{\mathbb{R}}^{dmodel}}\}_{i=1}^{Z}$ to producing a new set $\mathcal{Q}_{l+1}$ in the each layer. And for the target-points, we embedding them onto a higher ${{d}_{\operatorname{model}}}$-dimensional space by a linear layer without bias, i.e, ${{e}_{t}}=Linear(u)$. Since vehicle trajectory prediction involves predicting a sequence of future positions, it is important to include a position encoding that captures the temporal information of each past and future time step. More formally, the input embedding ${{e}_{w}}$ is time-stamped at time t by adding a positional encoding vector ${{p}^{t}}$ of the same dimensionality ${{d}_{\operatorname{model}}}$: $E={{p}^{t}}+{{e}_{w}}$.
Following\cite{Giuliari2020TransformerNF}, we use sine/cosine functions to define ${{p}^{t}}$.

\begin{gather}
 {{p}^{t}}=\{{{p}_{t,d}}\}_{d=1}^{{d}_{\operatorname{model}}} \nonumber \\ 
 \\
 where\quad {{p}_{t,d}}=\left\{ \begin{matrix}
   \sin (\frac{t}{{{10000}^{d/{d}_{\operatorname{model}}}}})\quad for\,\ d\ even  \\
   \cos (\frac{t}{{{10000}^{d/{d}_{\operatorname{model}}}}})\quad for\,\ d\ odd  \\
\end{matrix} \right. \nonumber
\end{gather}

And for the target-point embedding, we added a learnable encoding $U=u+{{p}^{u}}$.

Linear projections are employed in the Transformer architecture to calculate a group of queries, keys, and values (Q, K, and V). In order to compute attention between the waypoints and the target-point, our self-attention calculation is different from other Transformer structures:

\begin{gather}
 {{Q}_{a}}=E\centerdot M_{a}^{q} \nonumber \\
 {{K}_{a}}=Concat(E,U)\centerdot M_{a}^{k} \\
 {{V}_{a}}=Concat({{e}_{w}},u)\centerdot M_{a}^{v} \nonumber
\end{gather}


Where $M_{a}^{q}\in {{\mathbb{R}}^{Z\times {d}_{\operatorname{model}}}},M_{a}^{k}\in {{\mathbb{R}}^{(Z+1)\times {d}_{\operatorname{model}}}},M_{a}^{v}\in {{R}^{(Z+1)\times {d}_{\operatorname{model}}}}$ are weight matrices.

 Then the transformer utilizes the dot product of scaled query (Q) and key (K) matrices to compute the attention weights, and subsequently multiplies the attention weights with the corresponding value (V) matrix to obtain the final output.

\begin{equation}
\begin{aligned}
& SA=softmax (\frac{{{Q}_{a}}{{K}_{a}}^{T}}{\sqrt{{d}_{\operatorname{model}}}})V
\end{aligned}
\end{equation}
\textbf{Cross-Attention:}
Our cross-attention mechanism is similar to VIT\cite{Dosovitskiy2020AnII}. For the feature map output by Encoder $\mathsf{\mathcal{F}}\in {{\mathbb{R}}^{{{H}_{0}}\times {{W}_{0}}\times {d}_{\operatorname{model}}}}$, we reshape it into a
sequence of flattened 2D patches $memory\in {{\mathbb{R}}^{N\times {d}_{\operatorname{model}}}}$, where $N={{H}_{0}}\times {{W}_{0}}$ is the number of patches. And a learnable positional embedding is added to the patch to retain positional information.
 The calculation of Q, K, V matrices can be expressed by the following formula:

\begin{gather}
 {{Q}_{c}}=SA\centerdot M_{c}^{q} \nonumber \\
 {{K}_{c}}=memory\centerdot M_{c}^{k} \\
 {{V}_{c}}=memory\centerdot M_{c}^{v} \nonumber
\end{gather}


Then the cross-attention are computed according to Eq(6). Similar to the GRU-based method, our model predicts the waypoints offset rather than directly predicting the waypoints. Finally, we use the a FFN layer to back project the output of the decoder to the Cartesian person coordinates.

\textbf{Implement Details:}
Since the output dimension of Encoder is 512, we set the ${d}_{\operatorname{model}}$ to 512, and have 4 layers and 8 attention heads. For the activation function, because the network may input and output negative values, we tested different activation functions that can retain negative information and chose the best one: Leaky ReLU. Our FFN layer has 3 layers and also use the Leaky ReLU activation function. Our network was trained for 100 epochs using the AdamW optimizer with a learning rate of 0.0001. We applied a learning rate reduction of 0.1 at the 40th and 70th epochs.

\subsection{Loss Function}

we use the L2 loss between the predicted waypoint and the ground truth waypoint to train the network. For the time-step t, the loss function is given by:

\begin{equation}
\begin{aligned}
& \mathcal{L}=\sum\limits_{t=1}^{T}{{{\left\| {{w}_{t}}-w_{t}^{gt} \right\|}_{2}}}
\end{aligned}
\end{equation}


\begin{table*}[h]
\caption{Driving Performance Comparison}
\label{table1}
\centering
\begin{tabular}{lccc|ccc}
\toprule
\multirow{2}{*}{Method} & \multicolumn{3}{c|}{\textbf{Town05 Short}}          & \multicolumn{3}{c}{\textbf{Town05 Long}}            \\
\cmidrule(r){2-4} \cmidrule(l){5-7}
             & Driving Score$\uparrow$ & Route Completion$\uparrow$ & Infraction Score$\uparrow$ & Driving Score$\uparrow$ & Route Completion$\uparrow$ & Infraction Score$\uparrow$ \\
\midrule
\textbf{TAT-RT}(ours) & 44.68         & 79.44            & \textbf{0.59}    &  11.28        &  40.61           & \textbf{0.51}   \\
\textbf{TAT-CT}(ours) & \textbf{48.21} & \textbf{92.53}  & 0.53             & \textbf{11.31} & \textbf{69.08 } &  0.29           \\
\textbf{TAT-RR}(ours) & 10.26         & 67.43            & 0.15             &  5.23         &  40.24            &  0.14          \\
\textbf{TAT-CR}(ours) & 29.96         & 88.71            & 0.33             & 7.45          & 53.10            &  0.14           \\
\midrule
CILRS\cite{Codevilla2019ExploringTL}        & 7.44          & 13.56            & 0.49             & 2.12          & 8.12             & 0.14             \\
AIM\cite{Prakash2021MultiModalFT}          & 20.62         & 50.16            & 0.39             & 5.12          & 30.54            & 0.16             \\
Transfuser\cite{Prakash2021MultiModalFT}   & 43.31         & 82.97            & 0.55             & 9.68          & 66.66            & 0.19             \\
\bottomrule
\end{tabular}
\end{table*}

\begin{table*}[h]
\caption{Infractions Frequency Comparison}
\label{table2}
\centering
\begin{tabular}{lccccccc}
\toprule
\multirow{2}{*}{Method} & Driving score & \makecell[c]{Collisions with \\ pedestrians} & \makecell[c]{Collisions with \\ vehicles} & \makecell[c]{Collisions with \\ layout} & \makecell[c]{Red lights \\ infractions} & \makecell[c]{Off-road \\ infractions} & Agent blocked \\
\cmidrule(r){2-2} \cmidrule(l){3-8}
                        & \%$\uparrow$  & \#/Km$\downarrow$           & \#/Km$\downarrow$        & \#/Km$\downarrow$          & \#/Km$\downarrow$   & \#/Km$\downarrow$     & \#/Km$\downarrow$          \\
\midrule
\textbf{TAT-RT}(ours)            & 44.68         & \textbf{0.0 }                        & \textbf{1.97}                     
& \textbf{0.0 }                     &  27.17                   & \textbf{2.18}                 & 8.30             \\
\textbf{TAT-CT}(ours)            & \textbf{48.21}         & 0.48                        & 3.82                     & \textbf{0.0}                      &   26.75                  & 2.33                 & \textbf{1.43}              \\
Transfuser\cite{Prakash2021MultiModalFT}              & 43.31         & 1.63                         & 6.92                     & 2.66                      & \textbf{20.77}                  & 6.42                 & 4.79        \\
\bottomrule
\end{tabular}
\end{table*}

\section{Experiments}

\subsection{Task and Metrics}

We conducted all of our experiments on the CARLA simulator\cite{Dosovitskiy2017CARLAAO}, which provides a realistic urban driving environment. In this environment, the autonomous agent is required to follow a predefined route and navigate through various scenarios, including pedestrian crossings and obstacle avoidance, while receiving sparse goal locations in GPS coordinates (which we refer to as target-points) and discrete navigational commands.

To evaluate the performance of our method, we used the three metrics provided by the CARLA leaderboard: Driving Score (DS), Route Completion (RC), and Infraction Score (IS). RC represents the percentage of the route that the autonomous agent successfully completed, while IS measures the number of infractions made along the route, such as collisions with pedestrians, other vehicles, or traffic violations. The main metric, DS, is the product of RC and IS, and it provides an overall evaluation of the performance of the autonomous driving system.

\subsection{Dataset}

As CARLA provide 8 towns, we use 7 towns for training and hold out Town05 for evaluation. We generate routes randomly but the weather is fixed to ClearNoon. The routes are divided into Short routes of 100-500m and Long routes of 1000-2000m, and we ran a rule-based expert that can access privileged information to collect data at 2 FPS. The entire dataset contains 160k data.

\subsection{Baselines}
We compare our model with the following baselines. (1) CILRS\cite{Codevilla2019ExploringTL} predicts vehicle controls from a single front camera image while being conditioned on the navigational command. (2) AIM\cite{Prakash2021MultiModalFT} use  GRU-based waypoint prediction network with an image-based ResNet-34 encoder. It predicts future trajectory instead of vehicle controls, and it uses sparse goal locations as input instead of navigational commands. (3) Transfuser\cite{Prakash2021MultiModalFT} use multi-stage Transformer to fuse image and lidar features at multi scales, and a GRU-based auto-regression is used to generate future trajectory. It achieves SOTA on the CARLA leaderboard.

\section{Results}

\subsection{Comparison with baselines}

TABLE \ref{table1} presents the driving performance comparison between our method and the baselines we introduced earlier. Meanwhile, TABLE \ref{table2} shows the comparison between our model and Transfuser in terms of infractions frequency. We evaluated four variants of our model: \textbf{TAT-RT} uses Tranfuser as the backbone and auto-regressively predicts trajectory, \textbf{TAT-CT} uses Transfuser as the backbone and predicts trajectory as classification, \textbf{TAT-RR} uses ResNet34 as the backbone and auto-regressively predicts trajectory, and \textbf{TAT-CR} uses ResNet34 as the backbone and predicts trajectory as classification. The CIL-based method CILRS was found to be unsuitable for complex urban scenes, as it exhibited an extremely low driving score and route completion rate. We suspect that the main reason is the imbalance of the dataset. On the other hand, the ResNet network struggled to complete the perception task of autonomous driving, resulting in low driving scores for all methods. Nevertheless, our method outperformed AIM, with a 45.30\% improvement in driving score and a 76.85\% increase in route completion rate in the short route, and a 45.51\% improvement in driving score and a 73.87\% increase in route completion rate in the long route. It's important to note that a higher route completion rate indicates a higher chance of accidents, which can lead to a decline in driving score.

When compared to the original Transfuser model, our method increased the driving score by 11.31\% and the route completion rate by 11.52\% in the short route, and by 16.84\% and 3.63\%, respectively, in the long route. As for infractions frequency, our method reduced collision by 61.64\%, off-road infractions by 63.71\%, and the occurrence of agent blockages by 70.15\%. These results show that our method can make more effective use of perception features than GRU-based method, thus making more reasonable trajectory prediction. However, our method had a higher incidence of red light infractions. This could be due to the Transfuser perception model struggling to recognize traffic lights, as the original Transfuser model also had a high rate of running red lights. 

   \begin{figure*}[thpb]
      \centering
      \includegraphics[scale=0.5]{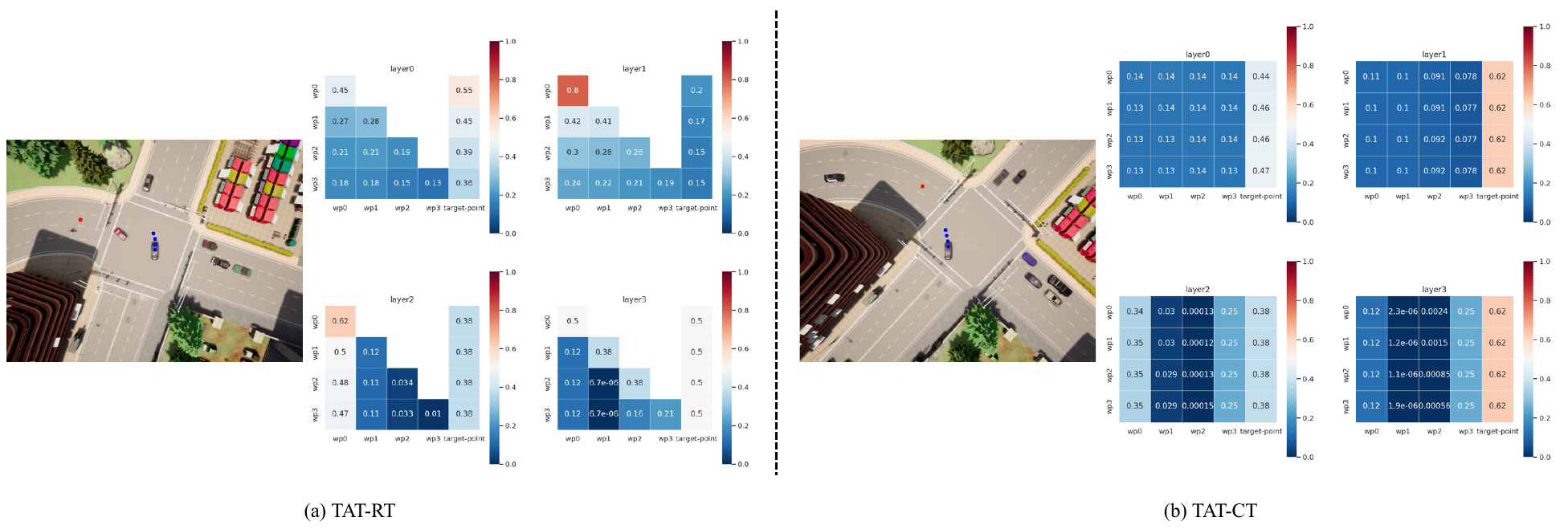}
      \caption{Self-Attention visualization of TAT-RT and TAT-CT, blue points are predicted waypoints, red points are target-points. The dependence of the auto-regressive method on the predicted waypoints leads to insufficient attention to the target-point, which may be the reason for the insufficient steering of the auto-regressive method}
      \label{figurelabel4}
   \end{figure*}

\subsection{Classification Vs Auto-regression}
In the field of trajectory forecasting, a recurring question is whether to approach the problem as an auto-regression or a classification task. The auto-regressive method generates future trajectory points one at a time, conditioning on the previously generated points, and is beneficial to the trajectory's continuity but has the problem of error accumulation. On the other hand, the classification method predicts all path points at once. In pedestrian trajectory prediction, the auto-regressive approach has been shown to be superior\cite{Giuliari2020TransformerNF}. However, in end-to-end autonomous driving, our research has reached the opposite conclusion.

In the Town05 short evaluation setting, compared to TAT-RT, TAT-CT has a 7.90\% higher driving score and a 16.48\% higher route completion rate, but has an 11.32\% lower infractions score. Specifically, the auto-regressive method can better avoid collisions (54.19\% lower collision rate), but the occurrence of agent blockages increased by 82.78\%. Figure \ref{figurelabel4} shows that insufficient steering is the main reason for the auto-regressive method's agent blocked, and this may be due to insufficient attention to the target point or error accumulation.

\subsection{Ablation study}
In our default configuration, we use Leaky ReLU activation function and add 1D sine/cosine positional embedding to waypoint queries and add learnable 2D positional embedding to memory. In this section, we present ablation on target-point attention architecture, activation function and the positional embedding, all within the context of the Town05 Short evaluation setting.

\textbf{Is target-point attention necessary?} 
To assess the effectiveness of the target-point attention structure, we compared it with an alternative approach that embeds the target point information into the waypoints queries (TET). However, we found that TET struggles to complete the route successfully. Notably, TET fails to navigate intersections accurately, suggesting that simply encoding target-points as position embeddings may not be sufficient to provide the network with enough target-point information. Our findings underscore the importance of the target-point attention structure for achieving successful end-to-end autonomous driving.

\textbf{Which activation function should be selected?} Since our model deals with negative values in both input and output, selecting the proper activation function is crucial. We evaluated three activation functions, namely tanh, Gelu, and Leaky ReLU, and observed that Leaky ReLU has the best performance. We found that the performance of Gelu significantly decreased, possibly due to less retention of negative value information. Meanwhile, the performance of tanh was slightly lower than that of Leaky ReLU, which might be caused by gradient disappearance and other issues.

\textbf{Is the positional embedding useful?} We expect that positional embedding can help the model understand the order between waypoints and the spatial dependency of the surrounding environment of the vehicle. TABLE \ref{table3} shows that positional embedding is indeed effective. No query PE or memory PE will result in significant performance degradation, and the performance of 2D memory PE is lower than that of learnable PE.

\begin{table}[]
\caption{Ablation Study}
\label{table3}
\centering
\begin{tabular}{cccc}
\toprule
Method           & \makecell[c]{Driving \\ Score$\uparrow$} & \makecell[c]{Route \\ Completion$\uparrow$} & \makecell[c]{Infraction \\ Score$\uparrow$} \\
\midrule
TAT(default)     & 48.21         & 92.53            & 0.53             \\
TET              & 21.33         & 30.96            & 0.70             \\
\midrule
TAT-gelu         & 36.51         & 64.60            & 0.59             \\
TAT-tanh         & 46.63         & 88.57            & 0.53             \\
\midrule
TAT-No Query PE  & 38.31         & 69.55            & 0.58             \\
TAT-No Memory PE & 36.52         & 64.60            & 0.60             \\
TAT-2D Memory PE & 37.89         & 75.92            & 0.55             \\
\bottomrule
\end{tabular}
\end{table}

\section{CONCLUSION}

In this paper, we proposed a novel trajectory prediction network for end-to-end autonomous driving. We demonstrate that the existing GRU-based trajectory prediction network fails to fully leverage the available perception features. To address this limitation, we propose a novel trajectory prediction network that leverages Transformer's attention mechanism to directly interact with high-dimensional perception features, and achieve state-of-the-art performance on CARLA. Our method is versatile and adaptable, and we plan to investigate further improvements by exploring new perception networks, such as separate traffic light detection networks to mitigate the issue of running red lights.





\section*{ACKNOWLEDGMENT}

This work was supported by the National Natural Science Foundation of China (U1964203), National Key Research and Development Program of China (No.2022YFB2503004), Sichuan Science and Technology Program (NO.2022YFG0342).


\bibliographystyle{plain}
\bibliography{citation}

\end{document}